%% file: main.tex
\documentclass[10pt,twocolumn,letterpaper]{article}

\usepackage[pagenumbers]{cvpr} 
\usepackage{soul}


\definecolor{cvprblue}{rgb}{0.21,0.49,0.74}
\usepackage[pagebackref,breaklinks,colorlinks,allcolors=cvprblue]{hyperref}

\def\paperID{697} 
\def\confName{CVPR}
\def\confYear{2025}

\title{RoomTour3D: Geometry-Aware Video-Instruction Tuning\\ for Embodied Navigation}

\vspace{-3mm}
\author{%
  Mingfei Han$^{1,3}$, Liang Ma$^1$,  Kamila Zhumakhanova$^1$,  Ekaterina Radionova$^1$,  Jingyi Zhang$^2$
  \\Xiaojun Chang$^{1,4}$,  Xiaodan Liang$^{1}$,  Ivan Laptev$^1$
  \\$^1$Department of Computer Vision, MBZUAI\quad
  $^2$Shenzhen Campus of Sun Yat-Sen University\\
  $^3$ReLER Lab, AAII, UTS\quad
  $^4$University of Science and Technology of China
  \\
  \url{https://roomtour3d.github.io}
}

\begin{document}
\maketitle
\input{sec/0_abstract}    
\input{sec/1_intro}

\input{sec/2_Related_Works}
\input{sec/3_RoomTour3D}
\input{sec/4_Vision-and-Language_Navigation_Model}
\input{sec/5_Tasks_and_Experiments}

\input{sec/6_Conclusion}
{
    \small
    \bibliographystyle{ieeenat_fullname}
    \bibliography{main}
}

\input{sec/X_suppl}

\end{document}

%% file: sec/0_abstract.tex
\begin{abstract}
\vspace{-4mm}

Vision-and-Language Navigation (VLN) suffers from the limited diversity and scale of training data, primarily constrained by the manual curation of existing simulators.
To address this, we introduce \textit{RoomTour3D}, a video-instruction dataset derived from web-based room tour videos that capture real-world indoor spaces and human walking demonstrations. 
Unlike existing VLN datasets, \textit{RoomTour3D} leverages the scale and diversity of online videos to generate open-ended human walking trajectories and open-world navigable instructions. 
To compensate for the lack of navigation data in online videos, we perform 3D reconstruction and obtain 3D trajectories of walking paths augmented with additional information on the room types, object locations and 3D shape of surrounding scenes. 
Our dataset includes $\sim$100K open-ended description-enriched trajectories with $\sim$200K instructions, and 17K action-enriched trajectories from 1847 room tour environments.
We demonstrate experimentally that \textit{RoomTour3D} enables significant improvements across multiple VLN tasks including CVDN, SOON, R2R, and REVERIE.
Moreover, \textit{RoomTour3D} facilitates the development of trainable zero-shot VLN agents, showcasing the potential and challenges of advancing towards open-world navigation.

\end{abstract}

%% file: sec/1_intro.tex
\vspace{-4mm}
\section{Introduction}
\vspace{-2mm}
Over the past years, Vision-and-Language Navigation (VLN)~\cite{reverie,r2r,wang2020environment,jain2019stay,liang2024cornav,ku2020room} has largely relied on human-designed simulators and annotated trajectories. R2R~\cite{r2r} established a benchmark for language-guided navigation in simulated indoor settings, while CVDN~\cite{cvdn}, REVERIE~\cite{reverie}, and SOON~\cite{soon} expanded VLN to dialogue-based and object-focused tasks. However, these manually curated simulations lack scene diversity and fail to capture real-world complexity.

To address limited diversity, recent methods propose the use of richer and more varied training data.
AirBERT~\cite{airbert} combines discrete Airbnb images for panoramic views, which lack consistency and naturalistic context of an indoor scene. 
ScaleVLN~\cite{scalevln} utilizes laboriously curated 3D scenes~\cite{xia2018gibson,ramakrishnan2021habitat}, but suffers from reconstruction quality and scalability. 
More recently, YTB-VLN~\cite{youtube_vln} attempts to use video frames to compose panoramic views and organize instructions with predefined templates, yet overlooks object variety and geometry structure.
NaVid~\cite{zhang2024navid} constructs sequential single-view trajectories from MatterPort3D~\cite{Matterport3D} and R2R~\cite{r2r} annotations, paired with general video data to train a sim-to-real agent.
None of these approaches simultaneously achieves scalability in scene diversity, openness in object variety, or comprehensive geo-perception in spatial representations, each of which is critical to training effective and open-world navigation agents.

To address the challenge, we introduce \textbf{RoomTour3D}, a novel dataset that provides a geometry-aware, spatially enriched training environment for VLN agents. 
Built upon easily accessible room tour videos from the Internet, RoomTour3D captures continuous movement through real estates with a hand-held camera from a first-person perspective.
Each frame presents a realistic, agent-centric view and showcases a rich array of indoor items. The continuous flow of these frames captures multiple views of the environment, presenting diverse room layouts and inherently embedding the geometric properties of the spaces.
To unleash the power of these videos, we propose an automatic and extendable pipeline to obtain open-ended 
geometry-aware human walking trajectories, spatially contextualized textual instructions using open vocabularies. 

To better model the navigation scenario, we take advantage of the continuous walk-through trajectories and densely sample frames from room tour videos. 
Then, we use COLMAP~\cite{schoenberger2016sfm,schoenberger2016mvs} to reconstruct 3D scenes of real-estates to obtain the geometric information.
With access to camera locations and orientations, we sample ``decision-making'' frames at points of maximal yaw rotation and sequence further frames every $\sim$1.5 meters to finalize trajectories. 
Additionally, our pipeline incorporates extensive annotations by employing RAM~\cite{zhang2023recognize} for object tagging, Grounding-DINO~\cite{liu2023grounding} for precise localization, and Depth-Anything~\cite{depthanything} to assess the relative distances between objects and the camera. 
To integrate knowledge of object variety, geometric awareness, and human walking preferences into model training, we employ GPT-4 \cite{openai2022gpt4} to generate navigation instructions for both summarization and task-specific navigation tasks.

Our RoomTour3D is an ongoing effort to create a comprehensive database derived from room tour videos and enriched by human-living knowledge. Currently, the dataset includes {\footnotesize $\sim$}100K open-ended trajectories with {\footnotesize $\sim$}200K descriptions, and {\footnotesize $\sim$}17K geometry-aware trajectories with navigable actions from 1847 homes. Moreover, we are releasing intermediate products such as object tags, bounding boxes, depth maps, room locations, and the necessary code and prompts used to generate instructions.
To validate the robustness, we conducted experiments with NaviLLM~\cite{navillm}, a generalist model based on Large Language Model (LLM), to train a unified multi-task navigation agent. Integrating our data into training simultaneously enhanced baseline performances such as CVDN, SOON, R2R and REVERIE with improvement exceeding 6\%, achieving an outstanding 9.8\% boost on SOON and setting new state-of-the-art (SOTA) results. 
Furthermore, our enriched action-instruction data enables the training of an end-to-end zero-shot navigation agent, advancing towards open-world embodied navigation.

In this work, we make the following key contributions:
\begin{itemize}
    \item \textbf{Video Collection for Complex Environments:} We curate a novel dataset of diverse videos tailored for navigation tasks, 
    distinguishing it from existing datasets such as YTB-VLN \cite{youtube_vln}. Our dataset features longer videos, enables representation of more complex environments, and exhibits fewer shot changes to ensure continuity and contextual consistency.
    \item \textbf{Automated Pre-processing of Videos:}
   We propose a pipeline to automatically extract geometry-aware navigation instructions, aligning spatial understanding with navigation goals. Additionally, we generate open-vocabulary instructions for diverse, open-ended trajectories to enhance real-world applicability.
   
   \item \textbf{Demonstrating Data Effectiveness:}
   Through extensive experiments and ablation studies, we demonstrate that our dataset significantly improves the performance of state-of-the-art models.
\end{itemize}



%% file: sec/2_Related_Works.tex
\vspace{-2mm}
\section{{Related Work}}
\vspace{-1mm}
\subsection{Vision-and-Language Navigation}
\vspace{-1mm}

\begin{figure*}[ht!]
\includegraphics[width=0.98\linewidth]{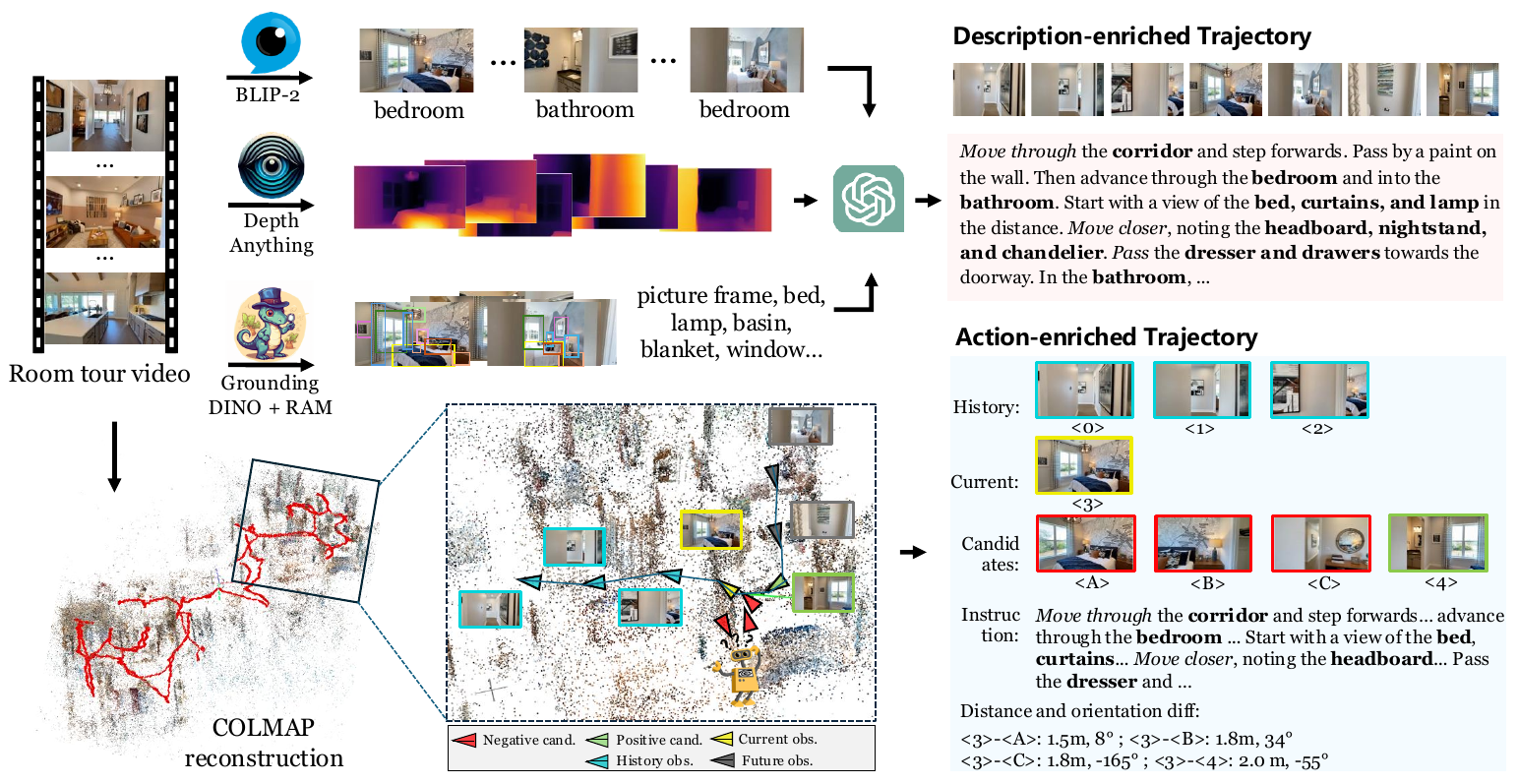}
\vspace{-1mm}
\caption{\small Overview of our RoomTour3D data generation. 
Starting from a room tour video, we first apply BLIP-2~\cite{li2023blip2} on frame sequence to predict the room locations. Next, we use RAM~\cite{zhang2023recognize} and Grounding-DINO~\cite{liu2023grounding} to identify objects within the frames and employ Depth-Anything~\cite{depthanything} for depth prediction. Subsequently, COLMAP is used to reconstruct the 3D scene with complete geometry information, and we sample human walking trajectories from the continuous frames.
The trajectory captures open-world objects, their positions, and depths relative to the camera. Finally, we use advanced LLM, \ie, GPT-4 to generate the free-form descriptions for pretraining, namely description-enriched trajectories. Specifically, 
for the trajectory shown in the figure,
which involves instant turning points, we specially treat \textless 0\textgreater\ to \textless 6\textgreater\ as walking trajectory, \textless A\textgreater\, \textless B\textgreater\, and \textless C\textgreater\ as side-watching points and use them as negative candidates for navigation finetuning task, namely action-enriched trajectories. For more details, please refer to Section~\ref{sec:data_generatopm}.}
\vspace{-3mm}
\label{fig:overall_pipeline}
\end{figure*}

Learning to navigate unseen indoor environments with natural language instructions is vital for enabling embodied agents to assist humans. Various scenarios have been explored, including fine-grained instruction following (R2R~\cite{r2r}) and dialogue-based navigation (CVDN~\cite{cvdn}), object localization from instructions (SOON~\cite{soon}, REVERIE~\cite{reverie}), and embodied question answering through active 3D exploration~\cite{eqa_1,eqa_2}. While substantial work focuses on task-specific models~\cite{hamt,duet,gao2023adaptive,kerm,hwang2023meta,vln_bert,long2023discuss,liu2024volumetric,Wang2024GOAT,zhao2024overnavelevatingiterativevisionandlanguage, metaexplore, soon, hao2020towards, qiao2022hop, li2023improving, qiao2023vln, an2023bevbert, hong20233dllm}, they often lack generalization across tasks. In this context, NaviLLM~\cite{navillm} introduces an embodied generalist model that simultaneously addresses multiple tasks through a single framework, demonstrating strong generalization ability.

\vspace{-1mm}
\subsection{Data-Centric Methods for VLN}
\vspace{-1mm}

The scarcity of VLN training data remains a critical issue and results in poor generalization of VLN agents to unseen environments. Most of existing VLN datasets such as R2R~\cite{r2r}, RxR~\cite{ku2020room}, CVDN~\cite{cvdn} and SOON~\cite{soon} are produced in simulators, which constrains data scalability due to the high labor costs involved. To tackle the problem, data augmentation~\cite{aug1,aug2,aug3,aug4,aug5,aug6,aug7} and self-exploration in simulator environments~\cite{self_explore1,self_explore2} have been investigated. 

VLN-BERT~\cite{vln_bert} and AirBERT~\cite{airbert} attempt to use web-based image-caption pairs for pre-training, however, resulting trajectories often fail to mimic realistic navigation. Similarly, automatic dataset generation pipelines~\cite{hm3d_auto,new_path}, including 
ScaleVLN~\cite{scalevln}, rely on manually curated 3D scenes or synthetic environments, which are costly to produce and lack the photorealism needed for robust real-world generalization. PanoGen~\cite{li2023panogen} enhances VLN training by generating diverse text-conditioned panoramic environments using text-to-image diffusion models and recursive outpainting. While this approach addresses the scarcity of training environments, it relies on synthetic panoramas and may not generalize well to real environments. 
YTB-VLN~\cite{youtube_vln} advances scalability by leveraging YouTube room tour videos to generate path-instruction pairs but omits explicit path geometry, essential for robotic navigation. 

In our work, we address the limitations by designing RoomTour3D with properties: (i)~free-form and open-vocabulary path annotations instead of template-based instructions, (ii)~extraction of open-ended trajectories from sequential video clips, and (iii)~inclusion of 
turning points and spatially close frames
as navigable candidate actions, moving beyond panoramic nodes. Furthermore, our approach integrates 3D reconstruction of indoor videos to retrieve trajectory geometry and employs an LLM to generate detailed, object-aware instructions with enhanced spatial understanding.

\vspace{-1mm}
\subsection{Zero-shot Navigation}
\vspace{-1mm}
Given the substantial semantic variations in complex real-world scenarios, fully-supervised VLN models often struggle to generalize across diverse navigation scenes. 
Zero-shot VLN has thus gained attention as it eliminates prior knowledge of environments and instructions, effectively mitigating environmental biases.

Commercial model based methods utilize advanced LLMs and robust frameworks for seamless solutions. MapGPT~\cite{chen2024mapgpt} incorporates a map-based prompting system with global spatial reasoning and adaptive path planning. DiscussNav~\cite{long2023discuss} employs a multi-expert framework where LLMs specialize in subtasks like instruction analysis and vision perception. NavGPT~\cite{zhou2024navgpt} focuses on explicit reasoning by combining commonsense reasoning with visual observations. As to non-commercial methods, LangNav~\cite{pan2023langnav} uses language as the primary perceptual space, while NavCoT~\cite{lin2024navcot} introduces parameter-efficient training to allow LLMs to autonomously reason and act.

We show that usage of our action-enriched data for navigation tuning results in superior zero-shot performance over all non-commercial methods and reaches comparable results to commercial approaches based on GPT-3.5.

%% file: sec/3_RoomTour3D.tex
\vspace{-2mm}
\section{RoomTour3D} 
\label{sec:data_generatopm}
\vspace{-1mm}


In this section, we present the automatic data curation pipeline of RoomTour3D. We detail the process of annotations, from sampling open-ended human walking trajectories to generating corresponding descriptions with open-world object variety and spatial awareness. Enabled by reconstructed 3D scenes, we further sample navigable trajectories with actions. The overall pipeline of our data generation is illustrated in Figure~\ref{fig:overall_pipeline}. Please refer to Appendix \ref{sup:vid_collection} for details about video collection.





\vspace{-1mm}
\subsection{Description-Enriched Trajectories}
\vspace{-1mm}

\begin{figure*}[ht]
\includegraphics[width=\linewidth]{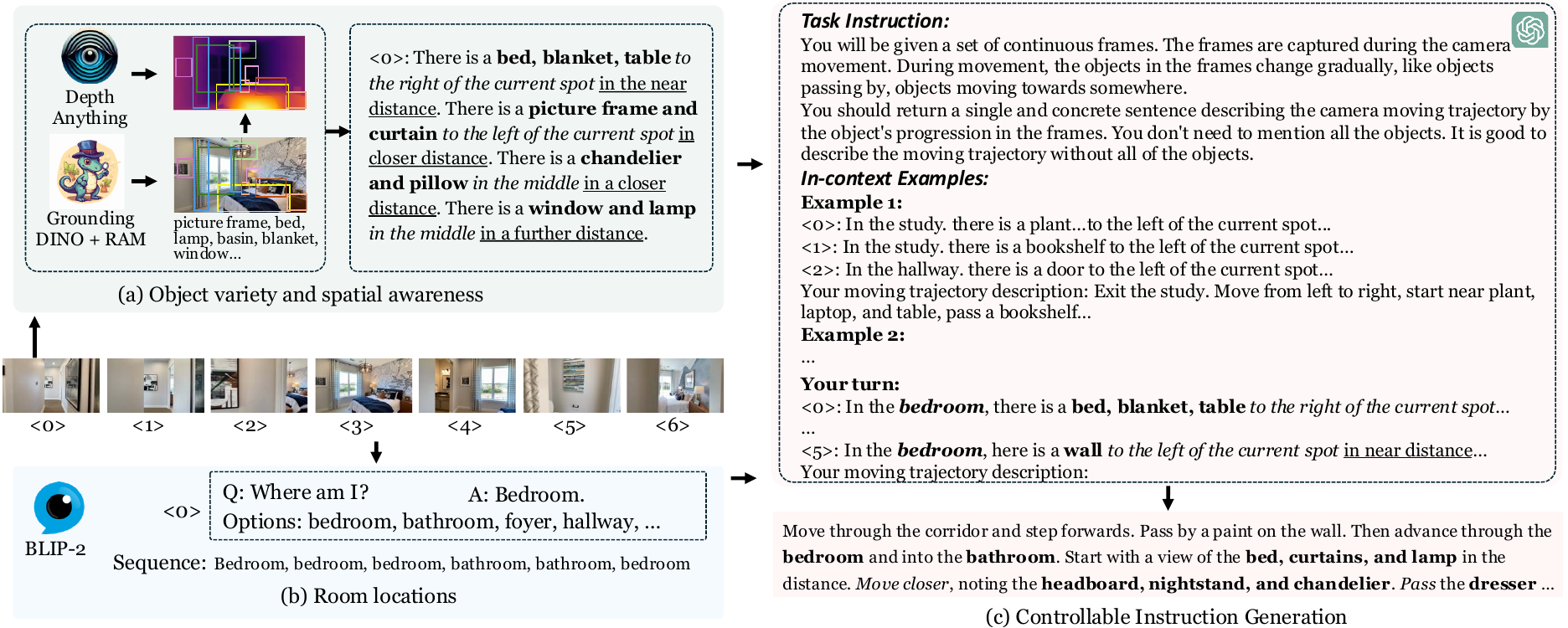}
\vspace{-4mm}
\caption{Instruction generation in a controllable way. (a) Using open-sourced expert models, we identify \textit{what objects are in the frames}, and assess \textit{how far an object is} and determine \textit{where an object is located}. The information is then textualized to create richly detailed frame captions. (b) BLIP-2 is adopted to predict and smooth room location across sequential frames. (c) Combining room locations and object information, we use GPT-4 for controllable and open-vocabulary instruction generation. The prompt consists of a task instruction that defines the generation task, and in-context examples that constrain the output style.}
\label{fig:generation_detail}
\vspace{-3mm}
\end{figure*}

In this subsection, we detail the process of generating controllable descriptions for open-ended trajectories. We start by generating human-walking trajectories by uniformly sampling frames at a rate of one frame every two seconds, which aligns with the average human walking speed of 1.42 meters per second~\cite{wikiwalking}, typically slower in indoor environments. Subsequently, to annotate these trajectories, as shown in Figure~\ref{fig:generation_detail}, we employ expert models such as BLIP-2~\cite{li2023blip2}, RAM~\cite{zhang2023recognize}, Grounding-DINO~\cite{liu2023grounding}, and Depth-Anything~\cite{depthanything} to gather extensive information on object variety, spatial positions, and depth measurements. Finally, we integrate this information into GPT4 \cite{openai2022gpt4} to generate detailed and coherent traje.

\noindent\textbf{Object Variety and Spatial Awareness.}
In order to harness object variety and enable spatial awareness, we compose three expert models and design a textual template, \ie, ``There is a \textit{object tag} to the \textit{spatial position} of current spot in \textit{relative distance}'', to organize the multi-source information to ease GPT generation.
Firstly, we used RAM~\cite{zhang2023recognize} to annotate the object categories within the frames. Based on these category tags, we employed Grounding DINO~\cite{liu2023grounding} to locate the objects in the frames. Subsequently, we used Depth-Anything~\cite{depthanything} to predict the depth maps corresponding to the frames. 


Using this data, we identify the spatial locations and distances of objects relative to the current camera position. By analyzing object bounding box center positions and depth map locations, we can generate frame captions, as illustrated in Figure~\ref{fig:generation_detail}. Finally, objects in different frames can be easily correlated and capture the progression across different frames. More details in the spatial awareness data generation are provided in Appendix \ref{supp_sec:instruction_generation}.

\noindent\textbf{Room Location Annotation in Videos.}
To determine the camera of each frame, w.r.t. the room category, we used BLIP2~\cite{li2023blip2} in visual question-answering mode, posing the question, "Which room am I in?" 
A predefined list of 16 common room types (e.g., bedroom, bathroom, kitchen) was used as possible answers. This list was curated by analyzing 10 randomly sampled long videos, using BLIP2 in generative mode to identify and rank the most frequently mentioned room types.
For frame-level predictions, we switched to BLIP2's discriminative mode and applied temporal smoothing to denoise outputs. 

We validated this approach by manually annotating 50 video clips, achieving an accuracy of 85\%. The use of BLIP2 leverages its open-world knowledge, while limiting room types to 16 categories for discriminative selection simplifies outputs and reduces ambiguity. Any loss in open-vocabulary flexibility is addressed during GPT-based trajectory summarization. 

\noindent\textbf{Controllable Instruction Generation.}
To generate descriptions that accurately capture human-walking trajectories and reflect the environment, we integrate frame-level room locations with frame captions composed of object descriptions. We then employ GPT-4-Turbo \cite{openai2022gpt4} for controllable instruction generation, leveraging the multi-source information contained in the composed captions. As depicted in Figure~\ref{fig:generation_detail}, we organize the prompt using the "Task instruction - In-context examples - Prediction" scheme. This approach defines our instruction generation task as describing object progression along the moving trajectories, and includes two examples to ensure GPT produces instruction-style texts only. As shown in Figure~\ref{fig:generation_detail} (c), captions of frames along the trajectory are embedded into the prompt and input into GPT.


\vspace{-2mm}
\subsection{Action-Enriched Trajectories}
\vspace{-1mm}


\begin{figure*}[t!]
\includegraphics[width=\linewidth]{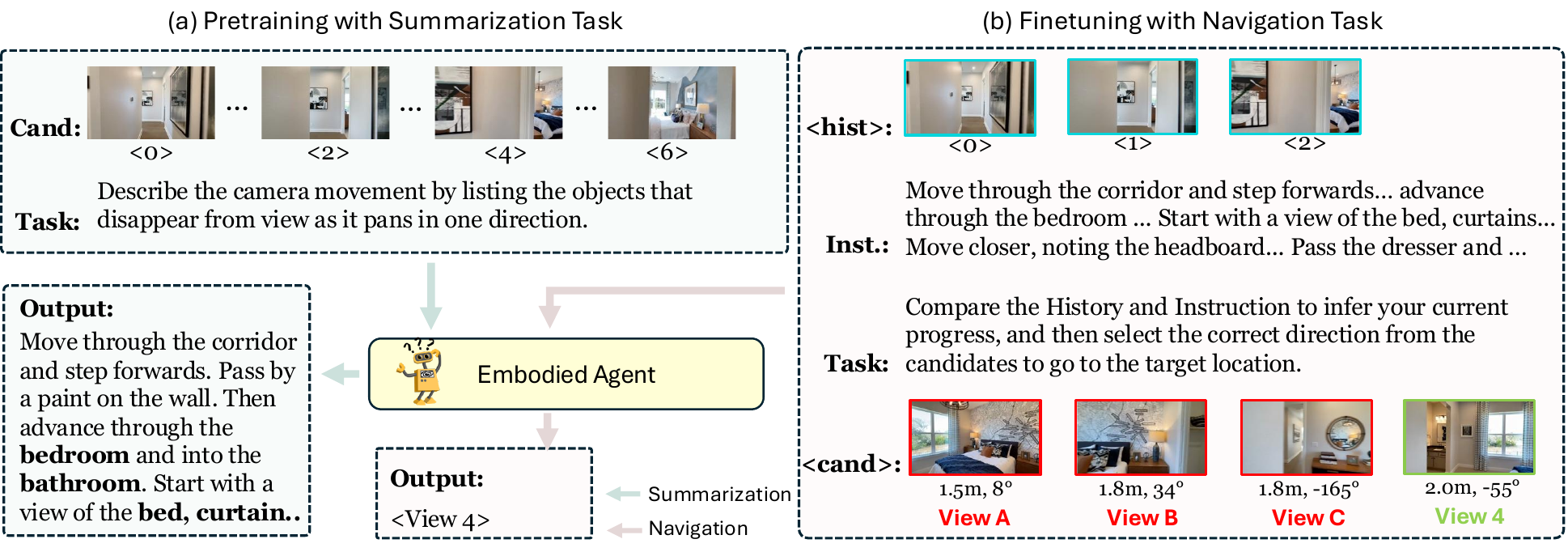}
\caption{Model training diagram with RoomTour3D. We design two tasks for our RoomTour3D to boost NaviLLM. (a) Pretraining: Sampled frames on the trajectory are treated as candidate observations. Model is optimized to summarize object progression along the path. (b) Finetuning: Each frame is considered as a navigable step. Given historical observation \textless 0\textgreater\ to  \textless 2\textgreater\ and navigation instruction, the model is prompted to predict the next action by selecting from candidate observations View A, View B, View C and View 4. }
\label{fig:method}
\vspace{-3mm}
\end{figure*}

\noindent\textbf{3D Environment Reconstruction.}
To obtain the geometric information of trajectories within RoomTour3D, we employ COLMAP~\cite{schoenberger2016sfm,schoenberger2016mvs} for 3D reconstruction. This process allows us to infer the 3D layout of environments in the videos, providing a detailed geometric context for navigation tasks. Specifically, we sample the videos at 3 frames per second to balance accuracy and execution time. To further improve time efficiency, we split the videos into 100-second video clips with 10-second overlaps between adjacent clips and perform reconstruction on the clips simultaneously.

Following this, we merge the resulting sub-models reconstructed from the video clips. For every two adjacent clips, if the reconstructed models have more than three overlapping frames, the models are merged and readjusted into one. However, due to varying reconstruction quality, a single video clip can produce more than one model. To manage this, we construct a graph for merging these sub-models. Each model is treated as a node, and we connect two models if they have more than three overlapping frames. We then apply Depth-First Search~\cite{depth_search} to merge any two connected models and replace the original model nodes with the new merged one, continuing this process until no connected nodes remain.




\noindent\textbf{Navigable Action Sampling.}
We enhance navigation action diversity by using frames from room tour videos as navigable actions by sampling at significant view-change points within a small radius. These points are identified by reconstructing 3D scenes to measure camera orientation differences and distances between frames, capturing varied views from revisited locations or turning points. Frames with substantial view changes are retained using cosine similarity thresholding, followed by non-maximum suppression to isolate major shifts. DBSCAN clustering \cite{dbscan} groups spatially close frames with different views to ensure diverse navigable actions.
These measures ensure the robustness to the misidentification of significant view change points that may arise due to inaccuracies in 3D reconstruction.
Finally, we identify distinct walking paths within each cluster. For each path, we select the most recent frame as a positive candidate and the frame with the highest angular difference as a negative candidate, enhancing the diversity of navigable actions. More detailed description of our navigable points sampling approach are provided in Appendix \ref{sup:nav_points}.

%% file: sec/4_Vision-and-Language_Navigation_Model.tex
\section{Vision-and-Language Navigation Model}

In this section, we introduce a practice to use our data to train a generalist embodied agent. To start with, we first provide a concise introduction to the state-of-the-art VLN model, NaviLLM~\cite{navillm}, which is an LLM-based navigation agent. Then, we introduce two tasks that are adapted for our RoomTour3D data, \ie, vision-instruction summarization task for pretraining and action-instruction navigation task for finetuning.

\subsection{Revisiting NaviLLM}

NaviLLM is a SOTA LLM-based model for embodied navigation, excelling on benchmarks like CVDN and SOON. It processes panoramic inputs by encoding environmental views and integrating them with navigation instructions.
Specific tokens are defined for different types of inputs: \textless hist\textgreater\ for historical observations and \textless cand\textgreater\ for candidate views at each navigational step.

During training, NaviLLM receives instructions and a sequence of candidate views. At each step, the candidate observations are input into the model along with the instructions. The model predicts the next action by selecting the appropriate view from the candidates, and this selected view is then cached as a \textless hist\textgreater\ token, updating the model's internal state for future decisions.
At the last step of the navigation task, the model summarizes all \textless hist\textgreater\ tokens as a separate training task to ensure comprehensive understanding and retention of the navigational history.
For testing, the model similarly uses historical observations, \ie, \textless hist\textgreater\ tokens, accumulated during navigation, and candidate views, \ie, \textless cand\textgreater, at each step to decide the next action. This process ensures that the model’s actions align with the given instructions and the observed environment.

\subsection{Summarization Task for Pretraining}

To leverage the rich, sequential nature of videos and enhance future planning capabilities, we adapt NaviLLM to use the 
description-enriched trajectories
from our RoomTour3D dataset for a summarization task. Each frame is treated as a candidate view and wrapped with \textless cand\textgreater\ tokens. Similar to selected panoramic views at each navigation step, these frames are considered as selected actions when executing navigation instructions.
As shown in Figure~\ref{fig:method}(a), the frame tokens and summarization task instruction are organized into a unified prompt and input into the LLM. The model is expected to output a trajectory summary containing object progression and room locations, as specified by the task instruction. The model is trained using next-token prediction loss, consistent with the original language model training methodology.





\subsection{Navigation Task for Finetuning}
In order to enable learning navigation decision-making from scalable scenes, we adapt NaviLLM to action-enriched trajectories from our RoomTour3D. Unlike panoramic views capturing observations from a single location, our data provides candidate views from frames at different locations and orientations, with only one frame directed toward the destination. Each frame in the video sequence is treated as a potential navigable action and wrapped with \textless cand\textgreater\ tokens. These candidate views are presented to the model and processed in the same way as panoramic views. As shown in Figure~\ref{fig:method}(b), the model processes the inputs to predict the next action, selecting the appropriate frame from the candidate views. The selected action is then cached as a \textless hist\textgreater\ token for subsequent decision-making steps.

During fine-tuning, each frame is treated as a navigable step, with the next trajectory frame as the target action and \textless STOP\textgreater\ as an alternative. The model uses historical observations and navigation instructions to iteratively predict the next action, building a detailed understanding of the path. At the final step, the model summarizes the navigation path, incorporating object progression and room locations. This summarization task enhances its ability to recall navigational history and improves performance.

%% file: sec/5_Tasks_and_Experiments.tex
\section{Tasks and Experiments}
This section outlines our experimental setup and presents the results. Detailed implementation information can be found in Appendix \ref{sup:impl_det}.

\begin{table*}[t!]
    \caption{Overall comparison with the baseline methods. Our RoomTour3D data can boost NaviLLM by a margin on SOON, R2R and REVERIE on SPL metric and on CVDN GP metric. $^\star$denotes reproduced results. RT3D$_{Desc}$ and RT3D$_{Action}$ stand for description-enriched trajectories only and action-enriched trajectories.}
    \vspace{-2mm}
    \label{tab:supervised_table}
    \centering
    \small 
    \setlength{\tabcolsep}{10pt}  
    \begin{tabular}{l|cccccccc}
    \toprule
    \textbf{Methods}            & \multicolumn{2}{c}{\textbf{CVDN}}          & \multicolumn{2}{c}{\textbf{SOON}}  & \multicolumn{2}{c}{\textbf{R2R}}               & \multicolumn{2}{c}{\textbf{REVERIE}}        \\
            & Val-U & Test       & Val-U & Test        & Val-U & Test     & Val-U & Test                 \\ 
    \midrule
    \multicolumn{9}{l}{\textit{Models Focusing on Single Task}} \\
    \midrule
    \multicolumn{1}{l|}{PREVALENT \cite{hao2020towards}}        & 3.15  & \multicolumn{1}{c|}{2.44} & -     & \multicolumn{1}{c|}{-}     & 53    & \multicolumn{1}{c|}{51} & -     & -                    \\
    \multicolumn{1}{l|}{HOP \cite{qiao2022hop}}              & 4.41  & \multicolumn{1}{c|}{3.24} & -     & \multicolumn{1}{c|}{-}     & 57    & \multicolumn{1}{c|}{59} & 26.1 & 24.3                \\
    \multicolumn{1}{l|}{HAMT \cite{hamt}}             & 5.13  & \multicolumn{1}{c|}{5.58} & -     & \multicolumn{1}{c|}{-}     & 61    & \multicolumn{1}{c|}{60} & 30.2  & 26.7                \\
    \multicolumn{1}{l|}{DUET \cite{duet}}             & -     & \multicolumn{1}{c|}{-}    & 22.6 & \multicolumn{1}{c|}{21.4} & 60    & \multicolumn{1}{c|}{58} & 33.7 & 36.0               \\
    \multicolumn{1}{l|}{VLN-SIG \cite{li2023improving}}          & 5.52  & \multicolumn{1}{c|}{5.83} & -     & \multicolumn{1}{c|}{-}     & 62    & \multicolumn{1}{c|}{60} & -     & -                    \\
    \multicolumn{1}{l|}{VLN-PETL \cite{qiao2023vln}}         & 5.69  & \multicolumn{1}{c|}{6.13} & -     & \multicolumn{1}{c|}{-}     & 60    & \multicolumn{1}{c|}{58} & 27.7 & 26.7                \\
    \multicolumn{1}{l|}{NavGPT2 \cite{zhou2025navgpt}}         & -     & \multicolumn{1}{c|}{-}    & -     & \multicolumn{1}{c|}{-}     & 61          & \multicolumn{1}{c|}{60} & - & -                \\
    \multicolumn{1}{l|}{BEV-BERT \cite{an2023bevbert}}         & -     & \multicolumn{1}{c|}{-}    & -     & \multicolumn{1}{c|}{-}     & \textbf{64}          & \multicolumn{1}{c|}{60} & 36.4 & \textbf{36.4}               \\
    
    \midrule
    
    \multicolumn{9}{l}{\textit{Unified Model For All Tasks}} \\ 
    \midrule
    
    \multicolumn{1}{l|}{NaviLLM(w. Pretrain)~\cite{navillm}}    & 6.16     & \multicolumn{1}{c|}{\textbf{7.90}}    & 29.2     & \multicolumn{1}{c|}{26.3}     & 59    & \multicolumn{1}{c|}{60} & 35.7 & 32.3     \\
    \multicolumn{1}{l|}{NaviLLM(w. Pretrain)$^\star$}    & 6.09    & \multicolumn{1}{c|}{-}         & 28.0     & \multicolumn{1}{c|}{-}          & 56.7     & \multicolumn{1}{c|}{-}       & 31.4     & -                    \\
    \multicolumn{1}{l|}{NaviLLM+RT3D$_{Desc}$(Ours)}      & \textbf{6.96}        & \multicolumn{1}{c|}{\underline{7.55}}        & \underline{30.2}         & \multicolumn{1}{c|}{\underline{26.5}}             & 62.3           & \multicolumn{1}{c|}{\underline{61.8}}     & 37.1        & 35.1               \\
    \multicolumn{1}{l|}{\textbf{NaviLLM+RT3D$_{Action}$(Ours)}} & \underline{6.33} & \multicolumn{1}{c|}{7.22}      & \textbf{31.7} & \multicolumn{1}{c|}{\textbf{27.8}}    & \underline{62.4 }& \multicolumn{1}{c|}{\textbf{62.2}}    & \textbf{37.4} & \textbf{36.4}
    \\
    \bottomrule
    \end{tabular}
    \vskip -0.05in
\end{table*}

\begin{table*}[h!]
    \vskip -0.05in
    
    \caption{Ablation study on the input modalities for trajectory summarization task.}
    \vspace{-2.5mm}
    \centering
    \small
    \setlength{\tabcolsep}{10pt}  
    \begin{tabular}{c c c |c| c c| c c | c c}
    \toprule
    \textbf{Object} & \textbf{Depth \& } & \textbf{Room} & {\textbf{CVDN}} & \multicolumn{2}{c|}{\textbf{SOON}} & \multicolumn{2}{c|}{\textbf{R2R}} & \multicolumn{2}{c}{\textbf{REVERIE}}  \\
    \textbf{tags} & \textbf{Bounding Box} & \textbf{type} & GP$\uparrow$  & SR$\uparrow$ & SPL$\uparrow$ & SR$\uparrow$ & SPL$\uparrow$ & SR$\uparrow$ & SPL$\uparrow$  \\
     
    \midrule
    $\times$ & $\times$ & $\times$ & 6.09 & 33.64 & 28.01 & 65.52 & 56.67 & 38.32 & 31.35 \\
    \checkmark  & $\times$ & $\times$ & 5.41 & 32.52 & 26.51 & 63.61 & 55.76 & 42.52 & 34.37 \\
    \checkmark & \checkmark & $\times$ & 6.49 & 37.62  & \textbf{30.40} & 68.37 & 61.70 & 41.72  & 36.04 \\
    \checkmark & \checkmark & \checkmark & \textbf{6.96} & \textbf{38.80} & 30.21 & \textbf{69.37} & \textbf{62.28} & \textbf{43.25} & \textbf{37.10}  \\

    \bottomrule
    \end{tabular}
    \label{tab:ablation}
    \vskip -0.1in
\end{table*}

\noindent\textbf{Datasets.}
During pretraining, we follow practice from NaviLLM~\cite{navillm} and perform teacher-forcing training on the combined dataset from our video-instruction data from RoomTour3D, together with CVDN~\cite{cvdn}, SOON~\cite{soon}, R2R~\cite{r2r}, REVERIE~\cite{reverie} and ScanQA~\cite{scanqa}, and augmented data from R2R and REVERIE. 
In the multi-task fine-tuning stage, we alternate between teacher forcing and student forcing on the combined data from our action-instruction data from RoomTour3D, together with CVDN, SOON, R2R, REVERIE, ScanQA and LLaVA-23k~\cite{liu2023llava}.


To evaluate the impact of our data on navigation agent training, we test on CVDN, SOON, R2R, and REVERIE. CVDN requires navigating towards a target by understanding dialog history, linking dialogue comprehension to actions. SOON tasks the agent with locating objects without bounding boxes, emphasizing semantic-visual alignment. R2R involves following step-by-step instructions, requiring dynamic progress tracking and precise alignment with navigational history. REVERIE focuses on localizing distant objects based on concise instructions, aided by ground truth bounding boxes at waypoints.

\begin{table}[ht]
    \vskip -0.02in
    \caption{Overall comparison with SOTA zero-shot methods on R2R. $^{\dagger}$ denotes training exclusive navigable actions. $^{\star}$ denotes using 36 views setting. RT3D stands for our RoomTour3D.}
    \centering
    \small 
    \begin{tabular}{l@{\hskip 18pt} | c c }
    
    \toprule
    \textbf{Methods} & \multicolumn{2}{c}{\textbf{Val Unseen}}  \\
                     & SR$\uparrow$ & SPL$\uparrow$ \\
    \midrule
    Random Walk~\cite{pan2023langnav}   & 3 &  2  \\
    \midrule
    \multicolumn{3}{l}{\textit{Commercial Model}} \\
    
    \midrule
    

    NavGPT(GPT-3.5)~\cite{zhou2024navgpt}$^{\star}$ & 13.89 & 9.12 \\
    NavGPT(GPT-4)~\cite{zhou2024navgpt} & 34 & 29 \\
    MapGPT(GPT-4)~\cite{chen2024mapgpt} & 38.8 & 25.8 \\
    MapGPT(GPT-4V)~\cite{chen2024mapgpt} & \textbf{43.7} & 34.8 \\
    DiscussNav(GPT-4)~\cite{long2023discuss} & 43 & \textbf{40} \\

    \midrule

    \multicolumn{3}{l}{\textit{Open-source Model}} \\
    \midrule
    LangNav(LLaMA2-7B)~\cite{pan2023langnav} & 0 & 0 \\
    NavCoT(LLaMA2-7B)~\cite{lin2024navcot} & 7.78 & 6.50 \\
    DuET (Init. LXMERT~\cite{tan2019lxmert}) & 1  & 0  \\
    NaviLLM~\cite{navillm}$^{\dagger}$ & 0 & 0 \\
    \textbf{NaviLLM+RT3D(Ours)}&~\textbf{14.33} & \textbf{10.86} \\
 
    \bottomrule
    \end{tabular}
    \label{tab:zero_shot}
    \vspace{-4mm}
\end{table}

\noindent\textbf{Evaluation Metrics.}
For the navigation tasks, we follow the evaluation methodology from \cite{r2r} using the following navigation metrics: \textbf{Success Rate (SR)}, which measures whether the agent reaches the target location within a set distance threshold; \textbf{Success Rate Weighted by Path Length (SPL)}, which is the SR adjusted by the ratio of the ground truth path length to the actual path traveled; \textbf{Goal Progress (GP)}, the advancement in meters towards the goal. GP is utilized for the CVDN dataset, whereas SR and SPL are the metrics for other datasets.

\subsection{Comparison on Supervised Tasks}

As shown in Table~\ref{tab:supervised_table}, we performed a one-time fine-tuning on the four tasks in a fully supervised manner. To begin, our experiments reiterate the superiority of multitask training over single-task training. Also, incorporating our RoomTour3D data into the pre-training process led to consistent improvements across all metrics on Val-U, achieving state-of-the-art results in the GP metric in the CVDN dataset. 

Notably, finetuning with our action-enriched data results in state-of-the-art performance on both Val-U and Test sets across SOON, R2R and REVERIE tasks. While the improvement on the CVDN and SOON datasets is modest, the most significant boost compared to the reproduced baseline is observed in R2R Val-U and REVERIE Val-U, with gains of approximately 5.7\% and 6\%, respectively. The improvement in R2R is largely driven by enhanced spatial awareness, stemming from the inclusion of proximity data, which helps the model better understand object distance and position. Similarly, gains in REVERIE are attributed to a combination of open-vocabulary tags, spatial awareness, and the addition of room type data, which encourages the model to infer the layout of environments, thereby boosting its spatial reasoning capabilities. Moreover, our use of open-ended instructions allows the model to adapt flexibly to diverse scenarios, fostering more robust and generalizable performance and better contextual understanding. 

\subsection{Comparison on Zero-shot Task}


\begin{figure*}[!ht]
    \centering
    \includegraphics[width=0.95\linewidth]{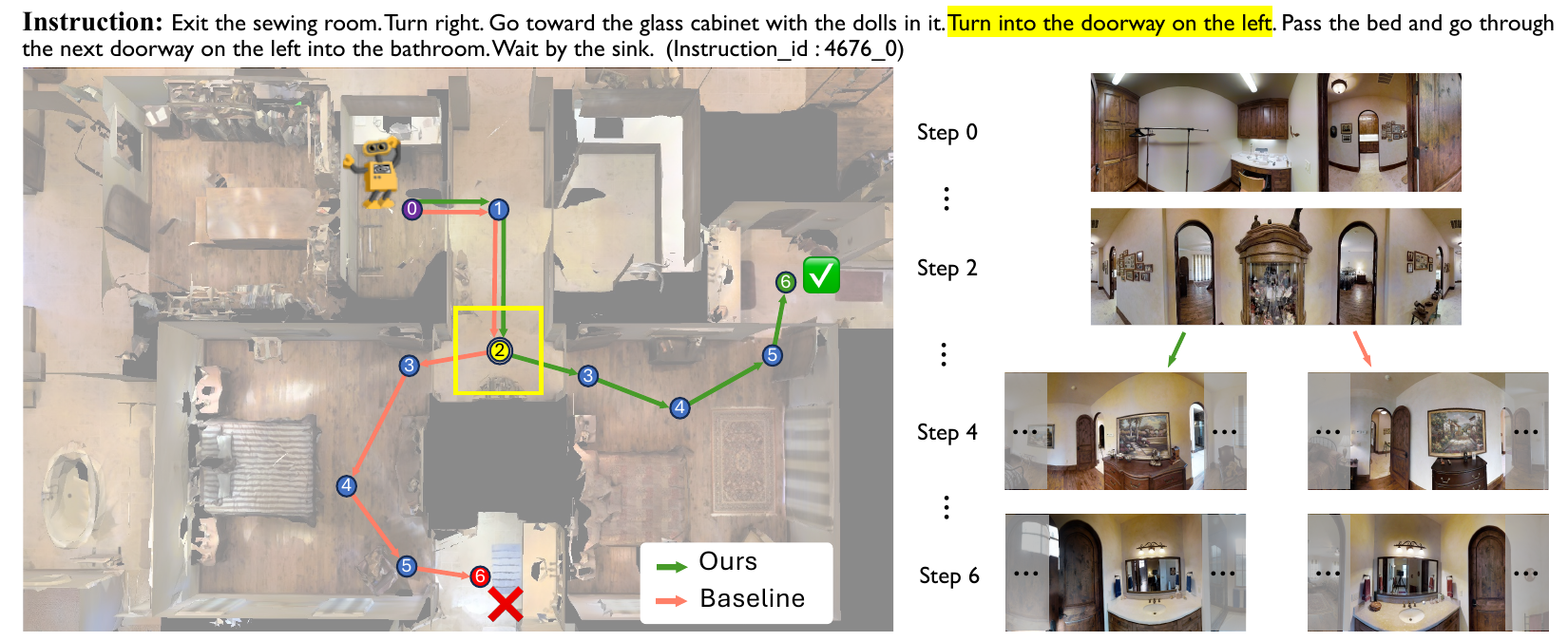}
    \vspace{-2mm}
    \caption{Paths of NaviLLM~\cite{navillm} and ours on R2R-unseen. Purple and green circles denote the start and target locations, respectively, and the red circle represents incorrect endpoint. According to the instruction, the agent should turn left at the waypoint marked with yellow. Our method makes the correct decision, while the baseline is confused by similar entrance at the waypoint, thus mistakenly turns right. }
    \vspace{-3mm}
    \label{fig:visualization}
\end{figure*}

To further demonstrate the substantial indoor knowledge contained in our data and its effectiveness for embodied action and language instructions, we conduct zero-shot experiments on embodied action prediction, as shown in Table~\ref{tab:zero_shot}. 

We removed all action and geometric data from the training datasets and retrained NaviLLM with and without our RoomTour3D dataset. Without action prediction data, NaviLLM lacked the ability to learn effective navigable action selection. However, with the inclusion of our action-enriched trajectories, NaviLLM achieved a 14.33\% SR and a 10.86\% SPL, outperforming open-source models built on LLaMA-7B and reaching results comparable to NavGPT~\cite{zhou2024navgpt}, which leverages GPT-3.5. These improvements validate the effectiveness of our 3D trajectories mined from room tour video reconstructions and emphasize the value of our action-enriched trajectories. This highlights the significant contribution of our dataset to advancing open-world navigation.

\subsection{Ablation study}

\noindent\textbf{Effect of open-world semantics and spatial awareness.}
As shown in Table~\ref{tab:ablation}, we analyzed the impact of various information types on instruction generation. Adding object variety significantly improved performance on REVERIE with increase SPL from 31.35\% to 34.37\%, as this dataset relies on object grounding. However, it had no direct impact on SOON, possibly because SOON relies solely on detailed textual descriptions without explicit bounding box annotations. After introducing depth estimation, which helps determine the relative distances of objects, the performance on SOON, R2R and REVERIE achieve marginal boosts. This demonstrates that enhancing spatial awareness significantly contributes to indoor navigation tasks.
Furthermore, incorporating room locations, which capture the scene semantics along the trajectory, provided a moderate boost across all four VLN tasks. This further highlights the critical role of object variety and spatial awareness in improving navigation performance.


\noindent\textbf{Effect of action-instruction data.}
As shown in Table~\ref{tab:supervised_table}, we test the effect of incorporating video-action-instruction data into the training dataset. It is evident that this approach improves the SPL metric across the test scenes of SOON, R2R, and REVERIE. We believe that incorporating geometric information and movement-inclusive instructions helps the model better align the relationship between action and observation changes, thereby further enhancing the model's embodied capabilities.

\subsection{Data correctness verification}
We evaluated the correctness of our automated data-generation pipeline by manually rating 100 randomly sampled trajectory descriptions on a 4-point relevance scale: 1 for ``totally irrelevant", 2 for ``partially relevant", 3 for ``mostly relevant" and 4 for a ``perfect match". The evaluation yielded an average score of 3.08, with 74\% of descriptions rated as "mostly relevant" or "perfect match," demonstrating the method's effectiveness in generating meaningful, visually aligned descriptions.


\subsection{Navigation Case Visualization}
As shown in Figure~\ref{fig:visualization}, selecting the correct action is critical at specific decision points, such as when a left turn is required to follow the instruction accurately. In the example, at step \textcircled{2}, both the rooms to the left and right could satisfy the latter part of the instruction, ``pass the bed and go into the bathroom.” However, the baseline method incorrectly chooses a right turn at the designated left-turn point, causing it to deviate from the intended path. Once this error occurs, even with scene graph history, the model struggles to realign with the correct trajectory. This challenge is particularly common in household environments, where bedroom layouts often appear similar. It further demonstrates the effectiveness of our data alignment in improving adherence to action-based instructions.

%% file: sec/6_Conclusion.tex
\section{Conclusion}
In this paper, we present RoomTour3D, a novel dataset automatically curated from room tour videos for VLN tasks. By leveraging the rich, sequential nature of video data and incorporating object variety and spatial awareness, we generate 200K navigation instructions and 17K action-enriched trajectories from 1847 room tour scenes. Additionally, we produce navigable trajectories from video frames and reconstructed 3D scenes, which significantly boost the performance and set new state-of-the-art results on the SOON and REVERIE benchmarks. This approach also enables the development of a trainable zero-shot navigation agent, demonstrating the effectiveness and scalability of RoomTour3D in advancing VLN research.

%% file: sec/X_suppl.tex
\clearpage
\appendix
\section*{Appendix}



The indexes of figures and tables in the appendix are continuous to the main sections for easy reference.

\noindent\textbf{Dataset release.} 
Our annotations and intermediate products are released at \url{https://huggingface.co/datasets/roomtour3d/roomtour3d} under CC-BY-SA-4.0 license. 
The downscaled and sampled video frames are released at \url{https://huggingface.co/datasets/roomtour3d/room_tour_video_3fps} under CC-BY-SA-4.0 license. 
The codes and project updates are hosted at \url{https://roomtour3d.github.io/}.

\noindent\textbf{Overview.} In the supplementary material, we provide
\begin{itemize}
   \item \textbf{Section \ref{sup:vid_collection}:}
   Room tour video collection process.
   \item \textbf{Section \ref{sup:nav_points}:}
   Navigable point extraction used for action-enriched trajectory generation.
   \item \textbf{Section \ref{supp_sec:instruction_generation}:}
   Object variety and spatial awareness for trajectory descriptions.
   \item \textbf{Section \ref{sup:room_reconstruction}:}
   Room tour 3D scene reconstruction.
   \item \textbf{Section \ref{sup:impl_det}:}
   Further model implementation details.
   \item \textbf{Section \ref{sup:qualitative}:}
   Qualitative results showcasing the instruction following capabilities of our trained model.
   \item \textbf{Section \ref{supp_sec:data_sample_vis}:}
   Data samples and excerpts from our data verification report to illustrate data curation correctness.
   \item \textbf{Section \ref{sup:impact}:}
   Broader impact of our work, including limitations and future extendable works.
   
\end{itemize}

\begin{figure*}[ht!]
    \centering
    \includegraphics[width=0.93\textwidth]{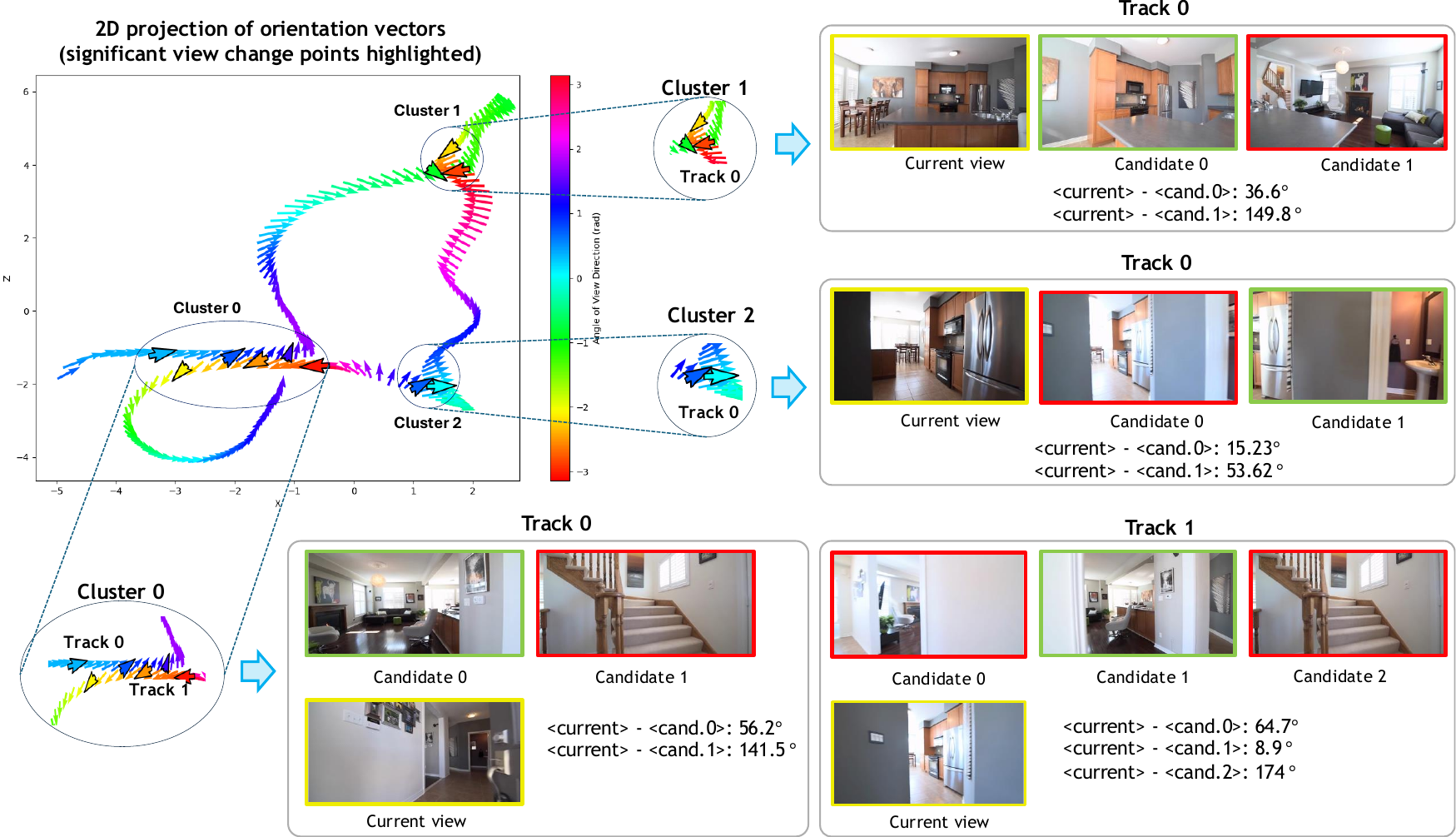}
    \caption{Visualization of significant view change point selection. For each cluster we identify the walking tracks and find the candidate views for the next action selection. This process ensures we have a diversified set of views in the setting without panorama images.}
    \label{fig:supp_turning_curve}
\end{figure*}

\section{Room Tour Video Collection}
\label{sup:vid_collection}
To enable more diversity for indoor scenes, we leveraged the rich variety and volume of room tour videos available on YouTube. These videos, recorded with hand-held cameras from a first-person perspective, offer a realistic and dynamic view of indoor environments. We curated a dataset from 1847 YouTube room tour videos, in total 243 hours. Our data collection approach builds on the video list from YTB-VLN~\cite{youtube_vln}, which we further filtered and expanded to enhance diversity and quality.

To ensure high-quality data, we prioritize continuous videos with least transitions, such as human interviews or abrupt cutting into close-ups, for better 3D reconstruction. We applied a title-description-based filtering process by using GPT-4~\cite{openai2022gpt4} and excluded videos shorter than three minutes. Additionally, we detected abrupt video transitions, retaining videos with at least nine continuous shots occupying over 80\% of the video duration. We further extended our dataset by continuously updating high-quality channels (\eg, NavaRealtyGroup, Open House 24, Sona Visual) with new videos, resulting in our current 1847 room tour scenes.

To process this data, we spatially downscale the resolution to shorter side 360 and temporally downsample the frame rate to 3 frames per second. All the following processing are performed on this downsampled data. 

\section{Navigable points generation}
\label{sup:nav_points}
To inject open-world knowledge from room tour videos into navigation agents, we propose navigating agents using video frames. Each frame in a human walking demonstration can be treated as having two next actions: move forward or stop. However, at significant view-change points — instances of distinct view shifts within a close radius — we sample frames with varied orientations as candidate actions to enhance the agent's training. Unlike YTB-VLN~\cite{youtube_vln}, which composes panoramic images at room nodes, our approach involves taking every significant view-change point and its neighboring frames that meet specific criteria as candidate actions.

First, we detect significant view-change points along person's trajectory. By reconstructing the 3D scene, we can determine the camera orientation difference and distance between frames. There are instances where the person may revisit a nearly identical location, resulting in varied views within almost the same spatial region. Additionally, turning points with notable view changes in close proximity are essential to capture. Identifying these view-change points is useful for producing diversified navigable action data, especially when panorama images are not available.

To find these points, for each point along the trajectory we calculated pairwise cosine similarity. We then applied a threshold of 45 degrees to retain only frames that demonstrate a substantial change in view. Afterwards, non-maximum suppression is performed along the trajectory to isolate local maxima in angular change to highlight the most significant view changes.

To account for the points that are close in proximity, but have different views due to an intersection in the walking trajectory, we performed DBSCAN clustering \cite{dbscan} of the points that were retained after Non-Maximum Suppression. This clustering step ensures a diverse set of navigable actions is maintained, even without the availability of panoramic images.

Finally, as shown in Figure~\ref{fig:supp_turning_curve}, to extract varied navigable action candidates, we post-processed the clusters by identifying the distinct walking paths of the person within each cluster.  In cases where paths intersect, the cluster may encompass two separate routes. For each walking path, we select the most recent frame on the walking path as a positive candidate, while a negative candidate is chosen as the frame within the cluster that exhibits the highest angular difference in view with the positive candidate.

\begin{figure*}[ht!]
    \centering
    \includegraphics[width=\textwidth]{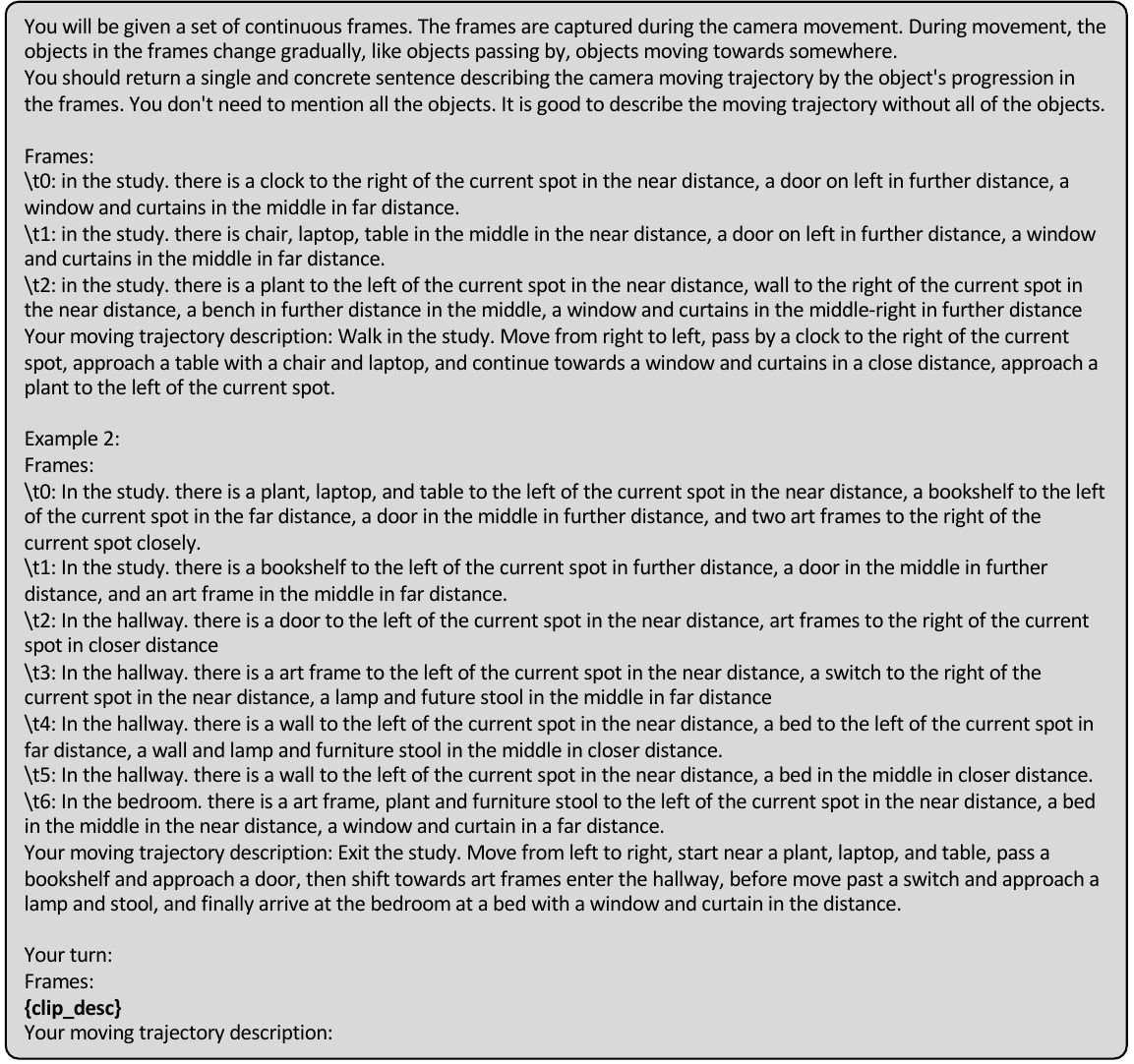}
    \caption{Prompt used for GPT-based instruction generation. We provide instruction for this generation task, in-context examples.}
    \label{fig:supp_gpt4_template}
    \vspace{-2mm}
\end{figure*}

\section{Instruction Generation}
\label{supp_sec:instruction_generation}

In this section, we detail the process of transforming spatial awareness and object variety information into textual captions for use with GPT. This involves extracting multi-source data using models such as RAM (Swin-L)~\cite{zhang2023recognize}, Grounding DINO~\cite{liu2023grounding}, and Depth-Anything~\cite{depthanything}, and then organizing this information into structured text inputs.

\begin{figure*}[ht!]
    \centering
    \includegraphics[width=\textwidth]{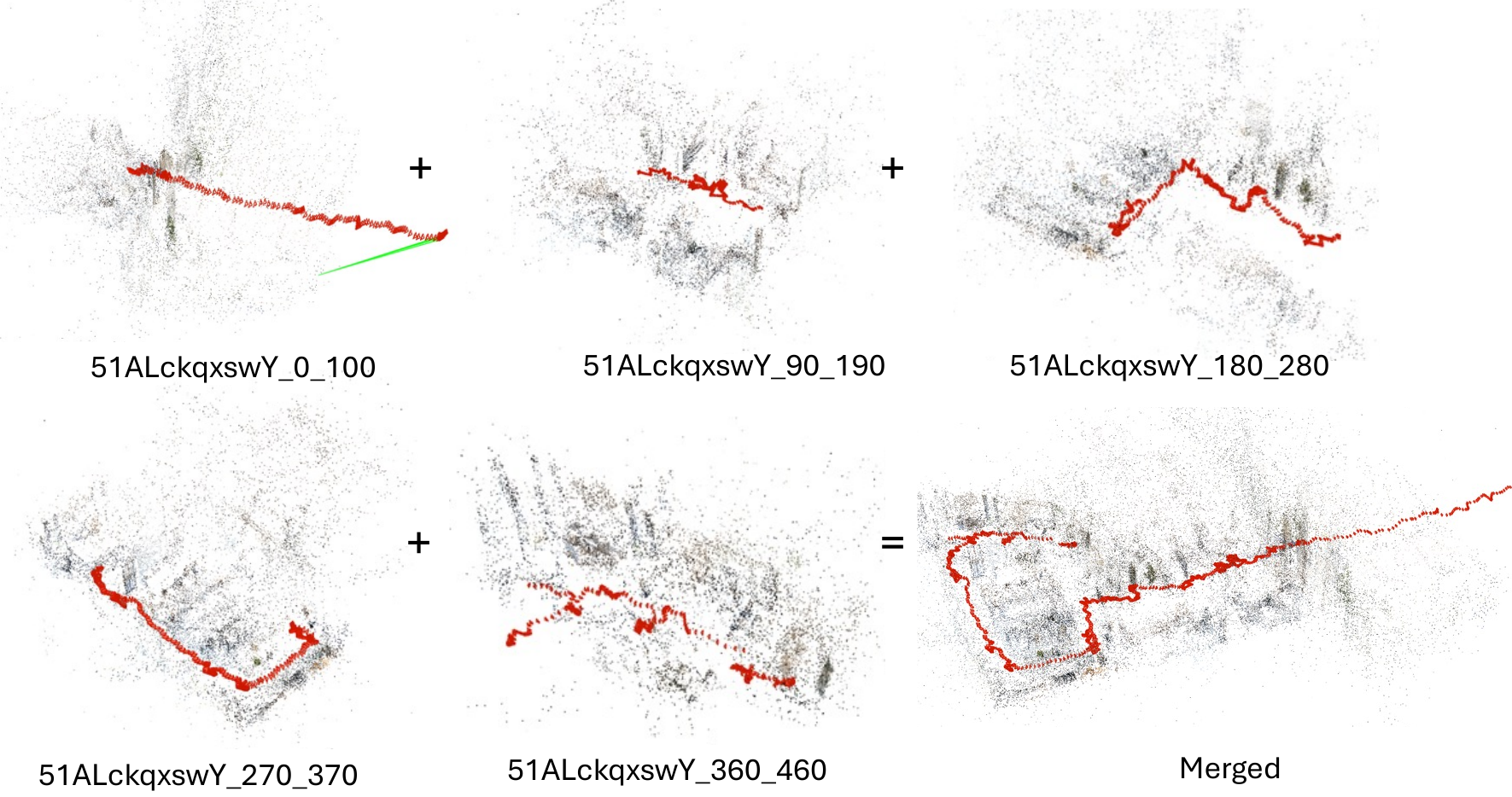}
    \caption{Illustration of the COLMAP model merging process. Reconstructed models from 5 adjacent video clips are successfully merged into one holistic model.}
    \label{fig:supp_colmap_merge}
\end{figure*}

\noindent\textbf{Object variety into texts.} 
Web videos offer a rich, open-world setup, capturing diverse items, arrangements, room functionalities, and layouts, which are critical for training open-world navigation agents. To fully utilize this diversity and ensure a controllable generation of instructions, we use RAM~\cite{zhang2023recognize} (Swin-L) to extract object tags in each frame. For each frame, we filter out the resulting entries indicating room types in order to be consistent with the identified room locations from BLIP-2. Then these object tags are used for grounding objects within the frames, for further integration of spatial awareness information.

\noindent\textbf{Spatial awareness into texts.} 
Navigation agents are frequently tasked with approaching or obtaining objects, making it crucial for them to sense object locations and dynamics during movement. To achieve this, we jointly use Grounding DINO~\cite{liu2023grounding} and Depth-Anything~\cite{depthanything} models to gather detailed spatial information. 
The reason why we used Depth-Anything over the depth derived from COLMAP~\cite{schoenberger2016sfm,schoenberger2016mvs} reconstructions is its ability to directly extract reliable depth without relying on long-range frames or structure-from-motion processes, which are prone to errors in complex video reconstructions.
This spatial awareness information is then transformed into text inputs suitable for GPT, enabling effective training.

We start by using Grounding DINO to spatially localize objects within the video frames. We define spatial locations relative to the capturing spot: \textit{to the left of the current spot}, \textit{in the middle}, and \textit{to the right of the current spot}. Specifically, the center 40\% of the frame is considered the middle, the leftmost 30\% as the left, and the rightmost 30\% as the right. For depth perception, we categorize distances into three ranges: \textit{in the near distance} (closest 30\%), \textit{in closer distance} (next 40\%), and \textit{in a further distance} (remaining 30\%).

Followingly, we integrate spatial location and depth estimation by measuring the overlap ratio between objects and the defined distance ranges. For example, if a carpet overlaps with the near-distance area by more than 30\%, we consider the carpet to be in the near distance to the capturing spot. Large objects that span multiple distance categories, such as a carpet visible in both near, closer, and further distances, are annotated accordingly to reflect their extended presence within the scene.

This structured approach ensures that our instructions capture the relative positioning and depth of objects, providing comprehensive context for navigation tasks. These texts are then further organized into GPT to generate contextually rich instructions for navigation agents training.

\noindent\textbf{GPT generation.} We utilize GPT-4 to summarize the object progression during the walking trajectory, leveraging the detailed object variety and spatial awareness texts. The template used for organizing the components is depicted in Figure~\ref{fig:supp_gpt4_template}. For each clip, we organize the object tags, spatial locations, relative distance to the camera and room locations per frame. This arranged content is then fed into GPT-4 to generate the trajectory summary and instructions. For data sample visualization, please refer to Sec.~\ref{supp_sec:data_sample_vis}.

\section{Room Reconstruction}
\label{sup:room_reconstruction}
To obtain complete geometric information, we adopt COLMAP~\cite{schoenberger2016sfm,schoenberger2016mvs} for indoor reconstruction. In this subsection, we detail the procedure of reconstructing room tour scenes, which further facilitates sampling navigable frames.

\noindent\textbf{Reconstruction of video clip.}
To reconstruct video clips, we start by sampling videos at 3 frames per second to balance accuracy and execution time. This frame rate provides sufficient detail for accurate 3D reconstruction while maintaining manageable processing times.
Each video is divided into 100-second clips with a 10-second overlap between adjacent clips.
Using COLMAP, we perform structure-from-motion and multi-view stereo processing on each clip. It estimates camera poses and generates a sparse 3D point cloud by identifying and matching feature points across frames. The command used for reconstruction is shown as follows, in which `\$DATASET\_PATH' denotes the folder containing sub-clip frames and reconstructed models will be located.
\begin{small}
\begin{verbatim}
colmap automatic_reconstructor \
    --image_path $DATASET_PATH/$IMG_FOLDER\
    --workspace_path $DATASET_PATH \
    --data_type individual \
    --quality high \
    --single_camera 1 \
    --sparse 1 \
    --dense 0 \
    --num_threads 10 --use_gpu 0
\end{verbatim}
\end{small}
\noindent\textbf{COLMAP model merging}
After performing individual reconstructions on video clips, we proceed to merge the resulting COLMAP models to create a unified 3D representation of the room tour scenes, as shown in Figure~\ref{fig:supp_colmap_merge}.

We begin by identifying overlapping frames between adjacent clips. These overlapping frames serve as common reference points for aligning and merging the separate models. If two reconstructed models share more than 3 common frames, we will try to merge these two models using the command as the following, where model merging and bundle adjustment are conducted in sequence.
\begin{small}
\begin{verbatim}
colmap model_merger \
    --input_path1 $MODEL_1 \
    --input_path2 $MODEL_2 \
    --output_path $RESULTED_MODEL_BEF_BA

colmap bundle_adjuster \
    --input_path $RESULTED_MODEL_BEF_BA \
    --output_path $RESULTED_MODEL_AFT_BA
\end{verbatim}
\end{small}
However, due to potential variances in reconstruction quality, a single video clip may produce multiple sub-models. To handle this, we adopt a graph-based approach for merging, \ie, Depth-First Search. In this approach, each sub-model is represented as a node in the graph. Edges are created between nodes that share more than three overlapping frames, indicating that these sub-models can be merged.

We iteratively merge the model nodes with edge connection existing by traversing from the first video clip (\eg, clip ``0\_100''). The successfully merged model will be a new graph node to replace the original separated two nodes.
In order to monitor the quality of this model merging operation, we use reprojection error to determine whether rolling back the merging operation. Specifically, if the error of the merged model is even larger than the sum of the two separate models, the model merging operation will be rolled back.
This iterative merging process continues until no further connections exist, resulting in a comprehensive and continuous 3D model of the room tour scenes. The final merged model provides detailed geometric information that is crucial for accurately sampling navigable frames and enhancing the training data for navigation agents.


\begin{figure*}[t!]
\centering
\includegraphics[width=0.9\textwidth]{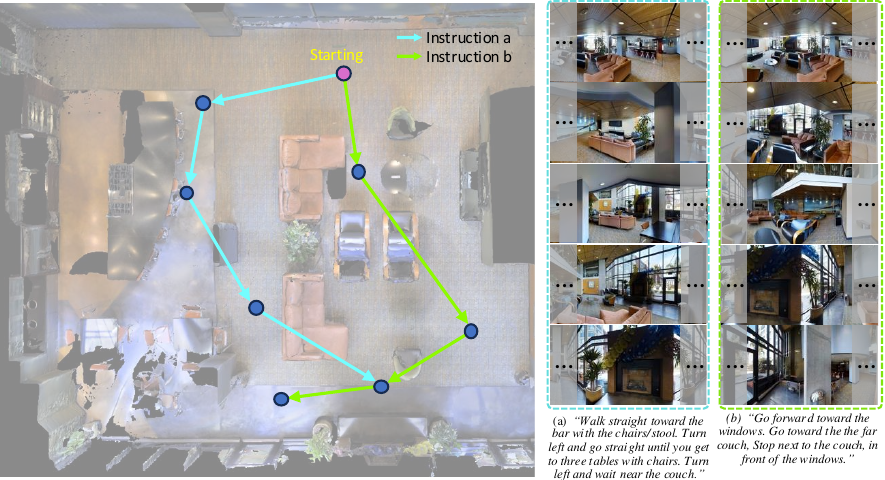}
\vspace{-1mm}
\caption{Visualization of the method trained with RoomTour3D on unseen scene \textit{8194nk5LbLH} with trajectory ID 4332. The agent successfully follows navigation instructions in R2R dataset. In (a), the agent first moves towards the bar and then approaches the couch. In (b), the agent moves forward towards the windows, then proceeds to the far sofa, and finally stops in front of the window.}
\label{fig:supp_vis_sample}
\vspace{-2mm}
\end{figure*}



\begin{figure*}[t!]
\centering
\includegraphics[width=0.95\textwidth]{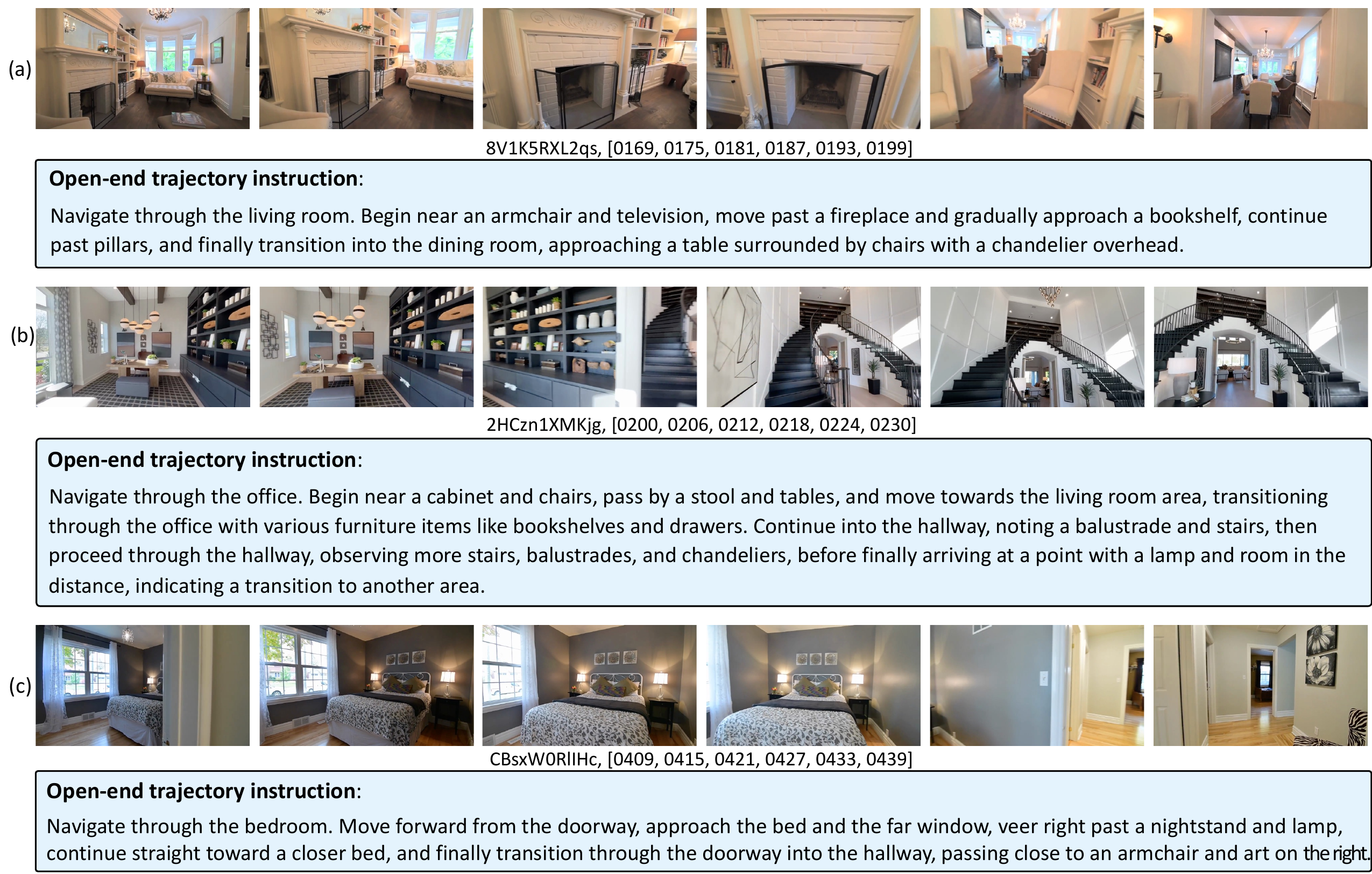}
\caption{Example open-ended trajectories and instructions. The instruction captures the surrounding environments and the object dynamics (``move past a fireplace'' in (a), ``move towards the living room area'' in (b)), and more importantly, the moving directions and destination (``approaching a table surrounded by chairs'' in (a), ``into the hallway, passing close to'' in (c)). All these data are automatically generated without manual correction.}
\label{fig:supp_data_sample}
\end{figure*}





\vspace{-2mm}
\section{Implementation details}
\vspace{-2mm}
\label{sup:impl_det}
Following the practice from NaviLLM~\cite{navillm}, we fine-tune the multi-view fusion module and the LLM. The multi-view fusion module consists of a 2-layer transformer encoder with a hidden size of 1024, and the LLM is built upon Vicuna-7B-v1.1~\cite{vicuna2023}. The ViT in the scene encoder is EVA-CLIP-02-Large, which remains frozen during training.
Our training follows a two-stage strategy using the Adam optimizer with a learning rate 3e-5. The model is trained for 2500 steps in the pre-training stage and 1250 steps in the multi-task fine-tuning stage, with a batch size 256. The training process utilizes 4×8 Nvidia A100 GPUs.
During testing, we employ a sampling strategy with a temperature of 0.01 for the SOON and REVERIE tasks to encourage exploration, while a greedy strategy is used for other tasks. This approach ensures robust performance across various evaluation scenarios.

\vspace{-2mm}
\section{Qualitative Results}
\label{sup:qualitative}
\vspace{-2mm}
This section presents qualitative results to demonstrate the effectiveness of our model trained with the RoomTour3D dataset. The model was evaluated on unseen scenes using the R2R dataset, focusing on its performance in following navigation instructions.
As shown in Figure~\ref{fig:supp_vis_sample}, we tested the model on an unseen scene, \textit{8194nk5LbLH}, with trajectory ID 4332. 
Experimented with two different instructions, the agent trained our data shows its flexibility in following the instructions. For example, in (a), the agent moves straight to the bar, then reaches the three tables with chars, and finally stops near the couch. In (b), the agent directly moves towards the window, following the instructions, then moves towards the far coach and stops. These results demonstrate the instruction-following navigation ability of the agent, which further highlights the effectiveness of our video-instruction data.

\section{Data Sample Visualization}
\label{supp_sec:data_sample_vis}
In this section, we present visualizations of data samples from the RoomTour3D dataset, as shown in Figure~\ref{fig:supp_data_sample}. These visualizations highlight the rich variety of indoor scenes, the spatial awareness embedded in the data, and the detailed annotations used for training navigation agents.

\begin{figure*}[ht!]
\centering
\includegraphics[width=\textwidth]{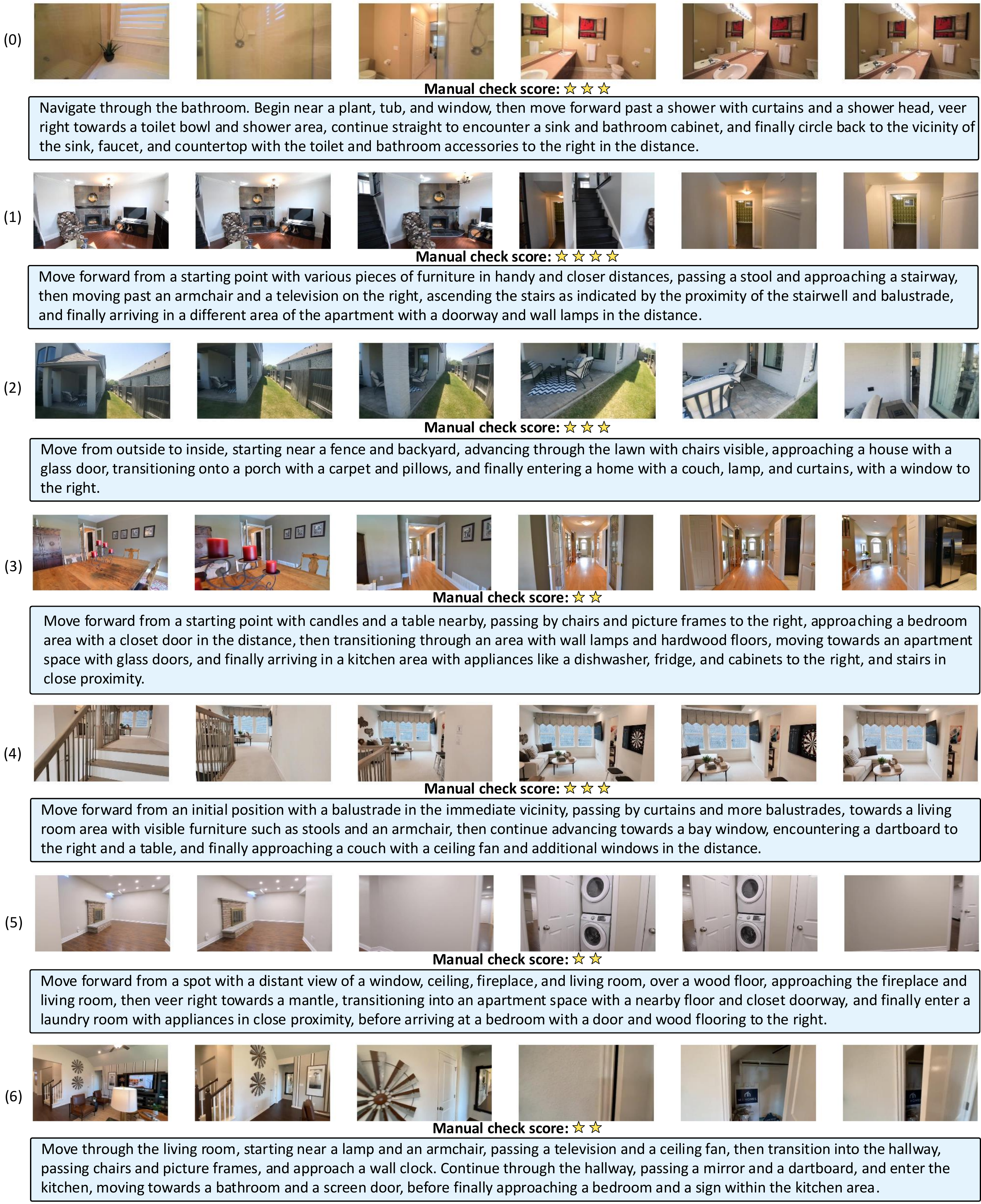}
\caption{Trajectory samples for manual check. For each trajectory, we provide frames and descriptions for check. The rating ranges from 1 to 4, representing ``totally irrelevant'', ``partially relevant'', ``mostly relevant'' and ``perfect match'' respectively. 7 out of 100 samples are shown here.}
\label{fig:supp_data_sample_manual_check_p1}
\end{figure*}
\begin{figure*}[ht!]
\centering
\includegraphics[width=\textwidth]{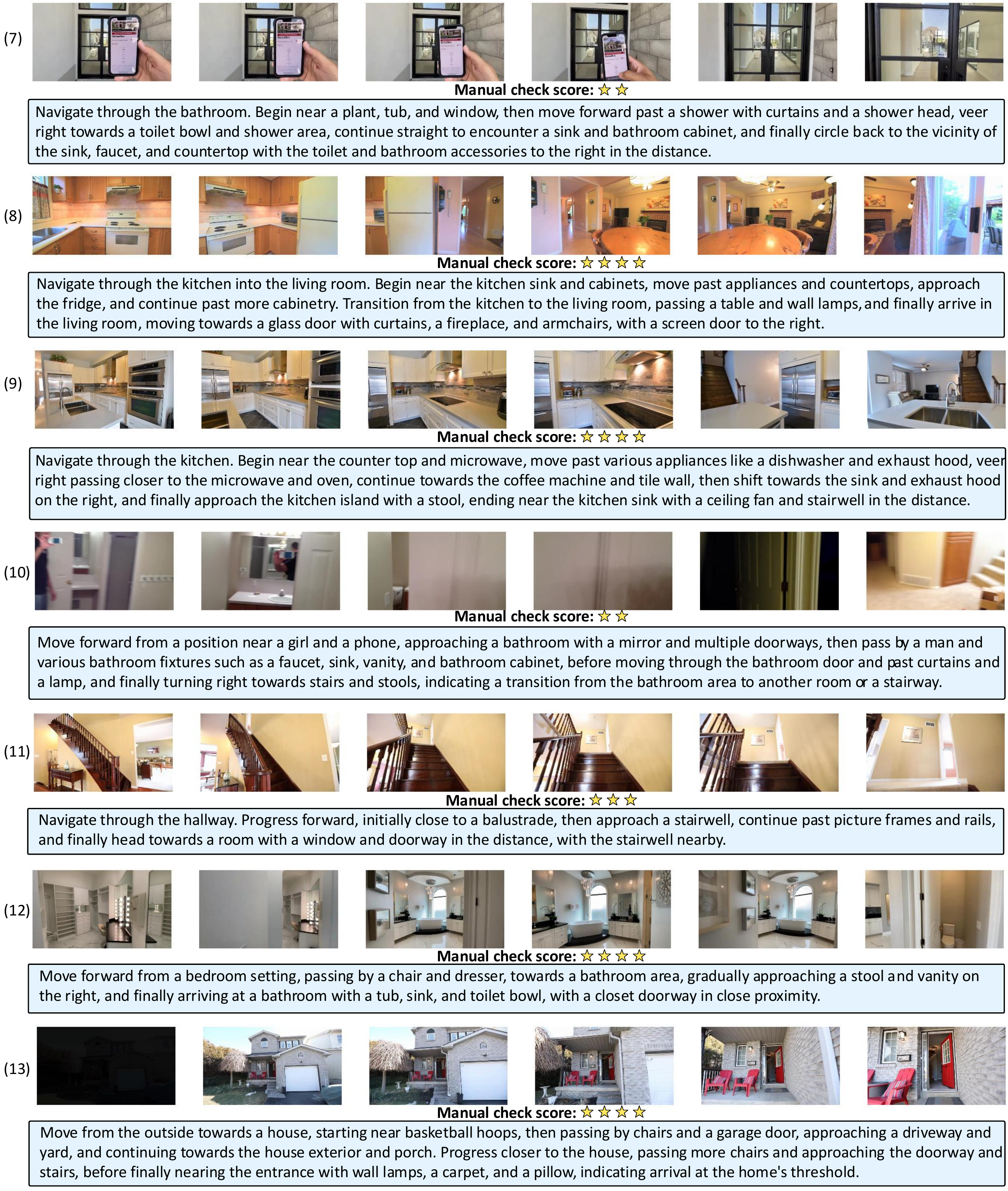}
\caption{Trajectory samples for manual check - Continued. For each trajectory, we provide frames and descriptions for check. The rating ranges from 1 to 4, representing ``totally irrelevant'', ``partially relevant'', ``mostly relevant'' and ``perfect match'' respectively. 7 out of 100 samples are shown here.}
\label{fig:supp_data_sample_manual_check_p2}
\end{figure*}

\noindent\textbf{Data correctness verification.} We provide part (14 out of 100) of manual check trajectories in Figure~\ref{fig:supp_data_sample_manual_check_p1} and Figure~\ref{fig:supp_data_sample_manual_check_p2}. For each trajectory, sampled frames and generated descriptions are provided, along with the manual check scores. The score ranges from 1 to 4, representing ``totally irrelevant'', ``partially relevant'', ``mostly relevant'' and ``perfect match'' respectively. Most of the sampled trajectories gain scores 3 and 4, which shows the convincing quality of our automatically generated descriptions.


\vspace{-2mm}
\section{Broader Impact}
\vspace{-2mm}
\label{sup:impact}
\noindent \textbf{Data Limitations and Ethical Considerations.} 
We provide downsampled video frames instead of the original videos. Users can also download these from the original sources. Additionally, our meticulous filtering process ensures that the video frames and annotations contain only indoor rooms and houses, containing no personally identifiable information or offensive content. The authors will take responsibility for long-term maintenance. 

\noindent \textbf{Scope of Conclusions.} It is important to recognize that experiments and data, including ours, might only represent a subset of universal realities. Nevertheless, given the wide range of room tour scenes covered in our videos, we believe our conclusions offer a robust understanding applicable to indoor embodied navigation. While specific to our dataset and results, these findings provide significant insight into the broader field of embodied navigation.

\noindent \textbf{Usage of Language Models and Simulators.} Our use of the LLaMA model\footnote{https://llama.meta.com/} from Meta, use of MatterPort3D data~\cite{Matterport3D} is authorized for research purposes. Those intending to use our model post-release should ensure they have the necessary permissions and adhere to usage restrictions. We express deep respect for the work of developers and contributors, recognizing their integral role in advancing language modeling and data collection.

\noindent \textbf{Future Research and Development.}
Aligned with our commitment to the research community, we released our code and dataset. This is intended to encourage further research and enable others to build upon our work. Although our current experiments require up to 8$\times$4 A100-80G GPUs for pretraining and 8 A100-80G for multi-task tuning, we are aware this may be a limitation.
Consequently, we plan to focus future efforts on adapting these experiments to be compatible with parameter-efficient tuned LLMs. It's important to note that fitting the experiments within an 8 GPU or fewer framework is not the primary focus of this paper. Still, we consider it a crucial step towards making our research more accessible and inclusive for various research groups.

Also, it would be interesting to investigate the usefulness of our data for grounded question-answering for 3D environments, particularly on the ScanQA dataset ~\cite{scanqa}.
